\documentclass[conference]{IEEEtran}
\IEEEoverridecommandlockouts
\usepackage{cite}
\usepackage{amsmath,amssymb,amsfonts}
\usepackage{algorithmic}
\usepackage{graphicx}
\usepackage{textcomp}
\usepackage{xcolor}
\usepackage{hyperref}

\def\BibTeX{{\rm B\kern-.05em{\sc i\kern-.025em b}\kern-.08em
    T\kern-.1667em\lower.7ex\hbox{E}\kern-.125emX}}

\usepackage{graphicx}
\usepackage{booktabs}
\usepackage{bbding}
\usepackage{wasysym}
\usepackage{gensymb}
\usepackage{tabularx}
\usepackage[linewidth=1pt]{mdframed}
\newtheorem{definition}{Definition}[section]

\begin{document}

\title{Spatio-Temporal Foundation Models: Vision, Challenges, and Opportunities}

\author{
\centering
\IEEEauthorblockN{Adam Goodge}
\IEEEauthorblockA{\textit{Institute for Infocomm Research (I\textsuperscript{2}R)} \\
\textit{A*STAR}\\ Singapore \\ goodge\_adam\_david@i2r.a-star.edu.sg}
\and
\IEEEauthorblockN{Wee Siong Ng}
\IEEEauthorblockA{\textit{Institute for Infocomm Research (I\textsuperscript{2}R)} \\
\textit{A*STAR}\\ Singapore \\ wsng@i2r.a-star.edu.sg}
\\
\IEEEauthorblockN{See-Kiong Ng}
\IEEEauthorblockA{\textit{Institute of Data Science} \\
\textit{National University of Singapore}\\ Singapore \\ seekiong@nus.edu.sg}
\and
\IEEEauthorblockN{Bryan Hooi}
\IEEEauthorblockA{\textit{School of Computing} \\
\textit{National University of Singapore}\\ Singapore \\ bhooi@comp.nus.edu.sg}
}

\maketitle

\begin{abstract}
Foundation models have revolutionized artificial intelligence, setting new benchmarks in performance and enabling transformative capabilities across a wide range of vision and language tasks. However, despite the prevalence of spatio-temporal data in critical domains such as transportation, public health, and environmental monitoring, spatio-temporal foundation models (STFMs) have not yet achieved comparable success. In this paper, we articulate a vision for the future of STFMs, outlining their essential characteristics and the generalization capabilities necessary for broad applicability. We critically assess the current state of research, identifying gaps relative to these ideal traits, and highlight key challenges that impede their progress. Finally, we explore potential opportunities and directions to advance research towards effective and broadly applicable STFMs.
\end{abstract}


\section{Introduction}\label{sec:intro}

\begin{figure*}
    \centering
    \includegraphics[width=0.75\textwidth]{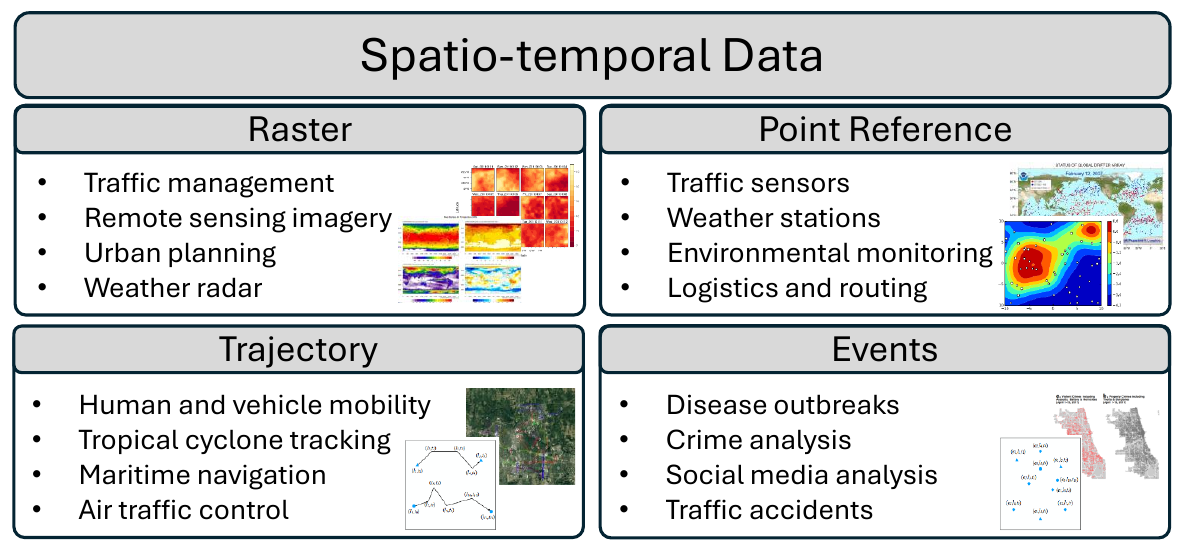}
    \caption{The four types of spatio-temporal data (raster, point reference, trajectory and events), with example use cases and illustrations.}
    \label{fig:data types}
\end{figure*}

The emergence of deep learning has significantly advanced state-of-the-art performance in an astonishingly diverse range of applications. In recent years, foundation models (FMs) \cite{bommasani2021opportunities}, which are large neural networks pre-trained on large-scale and broad data, have enabled transformative success due to their powerful abilities to generalize to a wide range of downstream tasks through the notion of transfer learning. Models like GPT \cite{brown2020language}, Llama \cite{touvron2023llama}, and BERT \cite{devlin2018bert} and DeepSeek \cite{guo2025deepseek} are trained on massive natural language corpora, enabling them to capture nuanced linguistic patterns which are important for various downstream tasks such as text classification, machine translation, summarization, and question answering. Similarly, models for computer vision such as CLIP \cite{radford2021learning}, ALIGN \cite{jia2021scaling} and SAM \cite{kirillov2023segment} have greatly improved performance across a wide variety of vision domains and tasks, such as classification, object detection and segmentation. There has also been very recent progress in foundation models for time-series data \cite{liang2024foundation} with models like Lag-llama \cite{rasul2023lag}, Time-LLM \cite{jin2023time} and Moirai \cite{woo2024unified}, where time-series tasks vary between domains and temporal frequencies. However, despite this success, FMs have yet to achieve a similar impact on performance in spatio-temporal tasks. Spatio-temporal (ST) data, encompassing data with both spatial and temporal dimensions, is pervasive across an extremely diverse range of fields, including urban analysis \cite{xie2020urban, pan2019urban, wang2022deep, yu2017spatio, tedjopurnomo2020survey}, weather forecasting \cite{ren2021deep, castro2021stconvs2s, ma2023histgnn}, climate science \cite{xu2021spatiotemporal, he2021sub, materia2024artificial, faghmous2014spatio}, environmental monitoring \cite{lin2018exploiting,abirami2021regional,amato2020novel,wen2019novel}, agriculture \cite{nevavuori2020crop, chen2023improving, croft2012use, xu2018spatio}, and public health \cite{song2021spatio,nikparvar2021spatio, yu2023spatio, wang2018predicting, zeng2021artificial}. Unlike time-series, ST data exhibit both spatial and temporal dependencies, which both need to be captured simultaneously. As the collection and distribution of spatio-temporal data continues to grow from diverse sources, so does the feasibility of spatio-temporal foundation models (STFMs) to learn shared patterns across different domains. This has great potential to improving efficiency, enhance generalization, particularly for data-deficient applications, and improve performance in a wide range of related tasks. However, progress remains slow due to the characteristics of ST data which makes training STFMs particularly challenging. Moreover, we observe that current research in this direction is highly fragmented between applications, which hinders progress towards a truly universal STFM comparable to those developed for other modalities.

In this paper, we present a vision for STFMs and aim to unify spatio-temporal foundation model research. We outline the key generalization capabilities that are essential for broadly applicable STFMs and examine the primary challenges and obstacles to their development. We critically evaluate the current state of research in this direction, identifying gaps relative to these ideals, and we explore opportunities for further advancement through targeted research and innovation. In summary, the key contributions of this work are as follows:

\begin{itemize}
    \item We propose a vision for the direction of spatio-temporal foundation models by identifying their key desirable capabilities.
    \item We examine existing efforts in STFM research and assess current capabilities with respect to these ideals.
    \item We consider the main avenues and opportunities for further research to further boost performance and applicability.
\end{itemize}

\section{Preliminaries}\label{sec:data}

\subsection{Spatio-Temporal Data Types}

Spatio-temporal data is any kind of data that involves both spatial and temporal dimensions. In practice, ST data can be organized into one of four main types or categories: \textbf{raster}, \textbf{point-reference}, \textbf{trajectory} and \textbf{event} data. Figure \ref{fig:data types} shows illustrations and example applications of each type of ST data, and we now describe them in more detail.

\textbf{Raster data} is ST data that is organized into a regular and fixed grid of cells, where each cell corresponds to a specific spatial location and contains a time-evolving sequence of measurements. With this, the data structure is represented as a four dimensional matrix: $X \in \mathbb{R}^{C \times L \times H \times W}$, where $C$ is the number of measured variables (or features), $L$ is the length of the sequence (i.e. number of time-steps), $H$ the height and $W$ the width of the grid. The total number of cells is $H \times W$. In practice, each cell may not correspond to a unique recording or sensor, as these are determined by physical or logistical constraints in the measured system. In this case, raw measurements are often processed into raster data of a desired resolution through various pooling and interpolation techniques. Due to its uniform structure and relative simplicity, raster data is widely used in various ST applications, ranging from transportation, weather \& climate analysis, medical imaging, remote sensing, and more. Video can be seen as a special case of raster data, where the measured variables are the RGB values in each pixel and each time-step corresponds to a frame. In this paper, we omit discussion of video data and their related applications, as their information-rich nature and complex pixel-level visual dynamics diverge significantly from the majority of spatio-temporal applications and are more appropriately aligned with computer vision.

\textbf{Point reference data} are ST sequences measured at specific point co-ordinates and may not neatly form a uniform grid. In this case, the data is represented as a three dimensional matrix: $X \in \mathbb{R}^{C \times L \times N}$, where $C$ represents the number of features, $L$ the sequence length and $N$ the number of points, which is typically (but not necessarily) equal to the number of sensors or  measuring devices. These positions may be fixed, for example weather stations which are fixed but not uniformly distributed through space. On the other hand, other devices such as weather balloons or sensors attached to ocean buoys are subject to atmospheric and ocean currents, meaning their position is subject the change from one time-stamp to another.

\begin{figure*}
    \centering
    \includegraphics[width=0.8\linewidth]{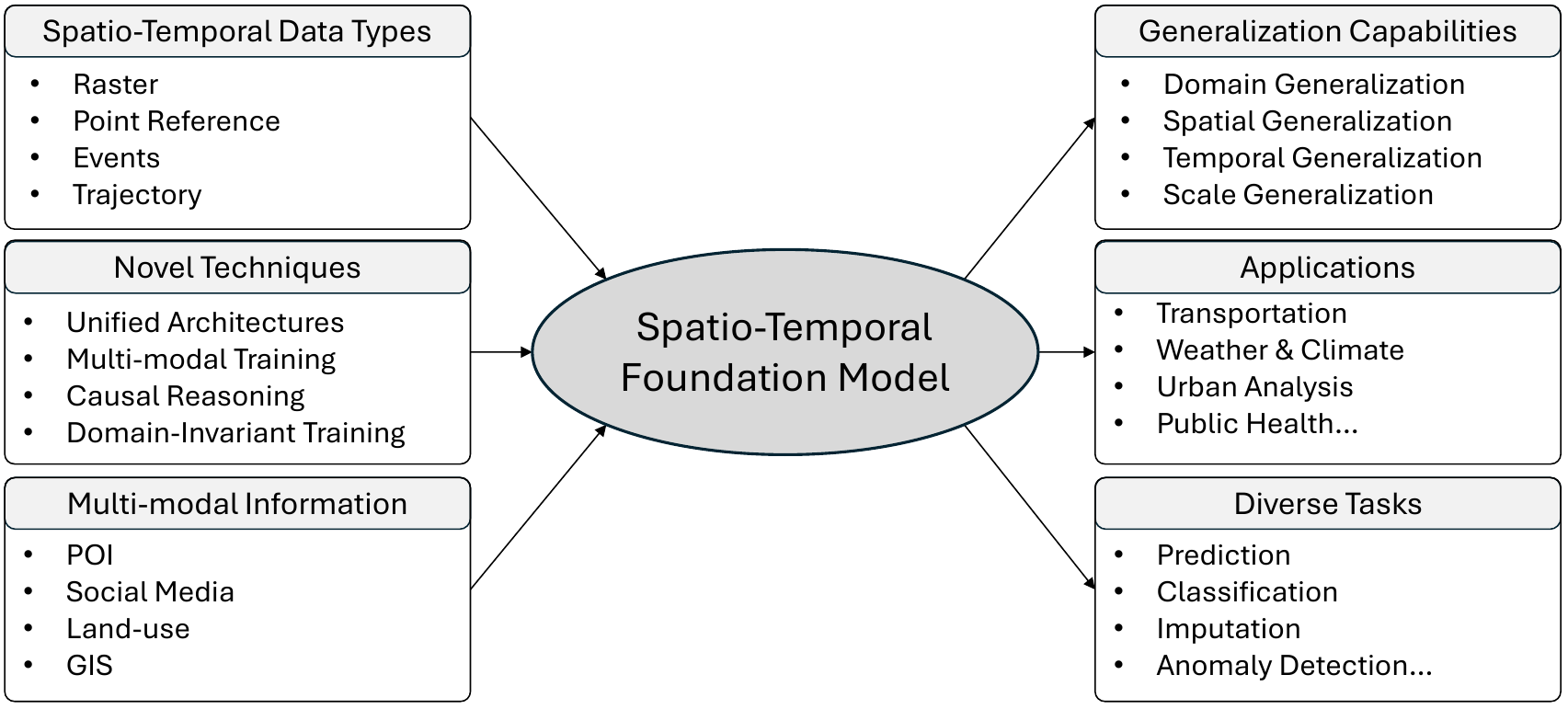}
    \caption{A framework for spatio-temporal foundation models (STFMs). Top-left: STFMs can flexibly handle various forms of ST data as input (see Section \ref{sec:data}). Middle-left: Novel techniques for training STFMs to handle multiple domains. Bottom-left: Various forms of complementary information in a variety of modalities can be incorporated into STFMs as guidance to perform specific tasks. Top-right: the generalization capabilities expected of STFMs (see Section \ref{sec:gc}. Middle-right: A variety of relevant applications for STFMs. Bottom-right: Diverse types of tasks that STFMs should perform.}
    \label{fig:framework}
\end{figure*}

\textbf{Trajectories} represent the paths traced by moving objects through space over time, consisting of paired sets of geographical co-ordinates and time-stamps: $\{l_i, t_i\}_{i=1:L}$. They are more suited to applications related to mobility, such as human, animal, vehicle or tropical cyclone tracking. In applications that involve several objects, it is common to discretize trajectories into buckets that represent a fixed spatial and temporal boundary, and the measured features indicate the number of trajectories found in each bucket. In this case, trajectory data is processed to more closely resemble raster data.

\textbf{Event Data} is characterized by a set of tuples $\{e_i, l_i, t_i\}_{i=1:N}$, where each tuple corresponds to a discrete event of type $e_i$, recorded at location $l_i$ and time $t_i$. Common applications include occurrences such as crime incidents, traffic accidents, disease outbreaks, natural disasters, and more. Event data can be particularly sparse, with large proportion of zero entries (i.e. no events within a given spatial and temporal interval), which presents unique challenges that require specialized techniques.

\cite{jin2023spatio} outlines two key properties shared by all kinds of spatio-temporal data, these are \textbf{heterogeneity} and \textbf{auto-correlation}. Heterogeneity means that ST patterns can vary across spatial (from one location to another) and temporal (from one time period to another) ranges and scales. This is a particularly challenging property for deep learning, as it is a violation of the i.i.d. assumption (that all data are independent drawn from an identical distribution), which is fundamental to the effective training of neural networks. Secondly, auto-correlation means that measurements taken closer together tend to follow more similar distributions, where closeness is understood in both the spatial and the temporal sense. This is neatly summarized in Tobler's First Law of Geography, which states that ``everything is related to everything else, but near things are more related than distant things".

\subsection{Spatio-Temporal Tasks}

\textbf{Prediction} is the task of forecasting future states or events based on historical patterns observed in spatio-temporal data. This process typically involves two components: the window, which represents the set of historical observations used as input, and the horizon, which defines the future time period to be predicted. In some cases, there can be a delay between the window and the horizon intervals, however it is more common to assume temporal continuity from the window to the horizon to maintain temporal continuity. Also, it is typical to assume that the window and the horizon contain an equal number of time steps. Prediction tasks can be categorized into short-term and long-term prediction, depending on the number of time steps involved, with short-term predictions focusing on immediate future states and long-term predictions concerning more distant outcomes, which are typically more challenging. Likely due to its practical importance in a variety of domains, spatio-temporal prediction has attracted the greatest amount of interest from the research community.

\textbf{Classification} tasks involve assigning labels to ST sequences out of a set of candidate labels. Examples include identifying land use from satellite imagery, classifying weather events, or vehicle types based on their trajectories.

\textbf{Clustering} aims to group spatio-temporal data points based on similarity. This is an important task for identifying and extracting relationships between different entities. Here, similarity can be flexibly defined to suit the application in question e.g. segmenting regions with similar temperature trends or identifying hotspots of disease outbreaks.

\textbf{Anomaly Detection} involves identifying spatio-temporal patterns that deviate significantly from the normal or expected patterns, such as detecting abnormal rainfall patterns or unusual movements in urban mobility. Anomalies are generally assumed to be rare in comparison to normal data, therefore anomaly detection methods often build a model of the expected behavior of a certain instance in order to detect any discrepancy to the norm.

\textbf{Change Detection} is the process of detecting significant deviations in spatio-temporal data over time. This is similar to anomaly detection, however anomalies are generally assumed to be isolated incidents, whereas change detection is aimed towards detecting sustained changes to the underlying environment, such as monitoring deforestation using satellite images or analyzing urban expansion. 

\textbf{Imputation} tasks focus on filling missing values in spatio-temporal datasets, such as gaps in weather station readings or satellite observations. Missing values can be temporal in nature, i.e. a missing time-step from an otherwise complete sequence, or spatial in nature, i.e. an entire sequence for an unobserved spatial location.

\subsection{Spatio-Temporal Data Mining}

\begin{figure*}
    \centering
    \includegraphics[width=14cm]{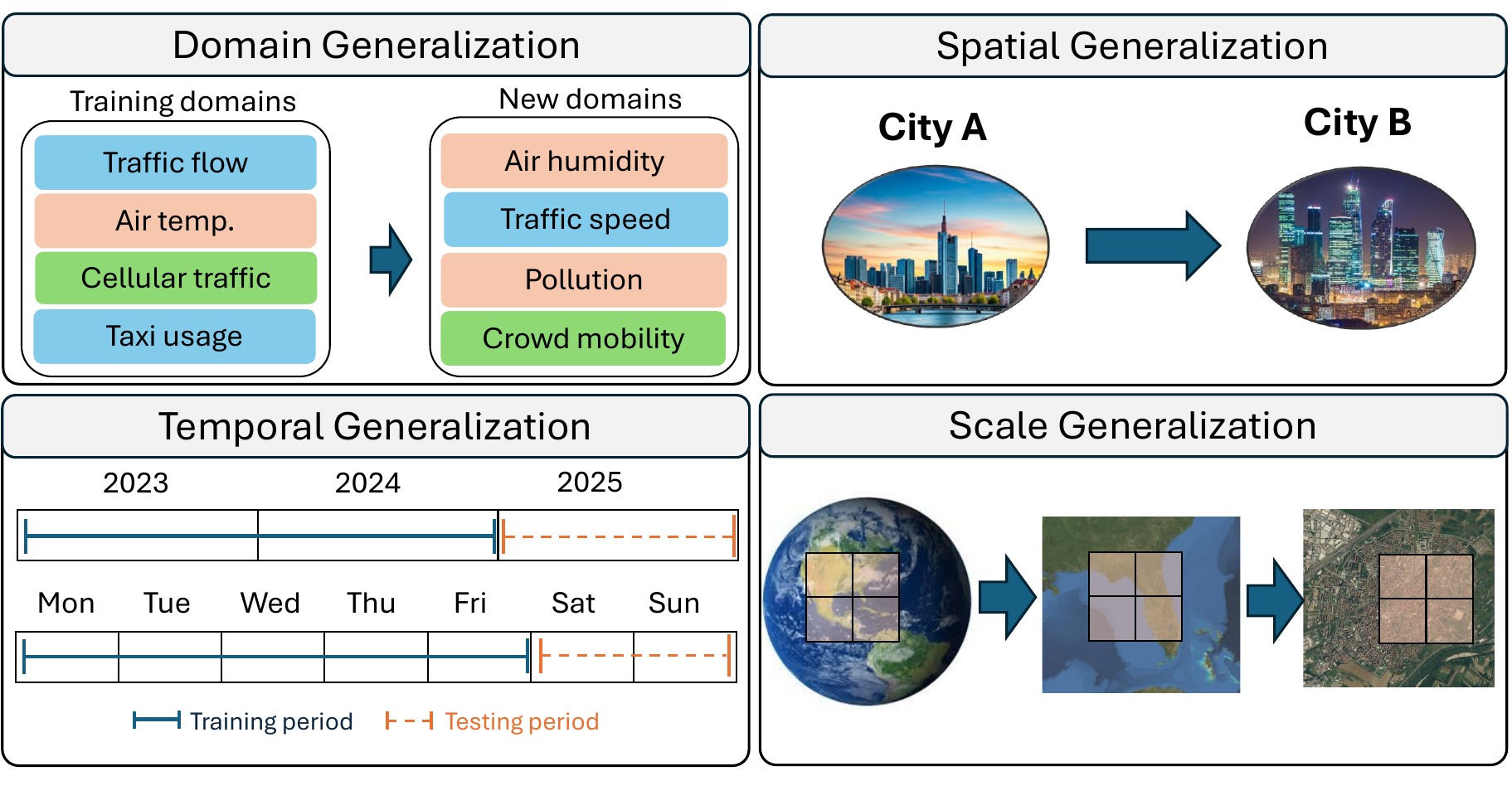}
    \caption{Four forms of generalization in spatio-temporal data. Top-left: domain generalization across different sources of data representing different physical systems and categories of applications. Top-right: spatial generalization: across different locations or regions in space. Bottom-left: temporal generalization across different periods and intervals of time. Bottom-right: scale generalization across different resolutions, frequencies or granularities of data.}
    \label{fig:enter-label}
\end{figure*}


Spatio-temporal data mining involves learning to model both the spatial and temporal patterns within ST data. A common approach has been to develop neural networks that integrate convolutional modules to capture spatial dependencies with recurrent modules to capture temporal dependencies. CNN-LSTMs use a convolutional neural network (CNN) to extract spatial features from input data, followed by a long short-term memory (LSTM) network to learn sequential patterns from the extracted spatial features \cite{bogaerts2020graph, yan2021multi, chen2019hybrid}. Alternatively, ConvLSTMs \cite{shi2015convolutional,lin2020self, azad2019bi, kim2017deeprain, wang2018deepstcl, diaconu2022understanding} replace the matrix multiplications in the LSTM gates with convolution operations to capture spatial dependencies within the sequential model.

In recent years, spatio-temporal graph neural networks (ST-GNNs) \cite{sahili2023spatio, kapoor2020examining, ta2022adaptive, mohamed2020social, yu2017spatio, ali2022exploiting, ma2023histgnn, zhang2020spatio, shao2022decoupled} have gained prominence due to their ability to flexibly handle ST data that does not conform to a regular grid structure. Instead, they operate over spatio-temporal graphs, which represent spatial positions as nodes or vertices in a graph, with connecting edges representing spatial relationships between adjacent nodes, such as proximity or connectivity. More details about ST-GNNs can be found in \cite{jin2023spatio}.

Following success in other modalities, the transformer architecture \cite{vaswani2017attention} has seen significant success in spatio-temporal applications \cite{yan2021learning, aksan2021spatio, yu2020spatio, li2022uniformer, grigsby2021long, luo2024lsttn} due to their ability to capture long-range dependencies across space and time through the self-attention mechanism. Unlike convolutional models, which operate over local receptive fields, or recurrent models, which depend on sequential processing, transformers can learn global relationships by attending to all parts of an input sequence simultaneously. This is particularly useful in domains where complex, non-linear interactions evolve over large spatial areas and over long temporal periods. Transformers are also the architecture of choice for foundation models for both image and language modalities, though the vast majority of ST research continues to adopt a one-model-per-task paradigm which requires training separate models for individual tasks.

\section{Spatio-Temporal Foundation Models}

The key distinction between the one-model-per-task-paradigm and the foundation model paradigm is their notion of \textbf{\textit{generalization}}. 

\begin{definition}
    \textit{Generalization is the ability of a model to learn patterns which effectively transfer from one set of data to another.}
\end{definition}

In the one-model-per-task paradigm, models are trained to perform a single task with data from a single domain and only expected to generalize to unseen samples from the same underlying distribution. On the other hand, foundation models are trained with much broader data and expected to generalise to new and unseen data drawn from other distributions. With this context, we define a spatio-temporal foundation model as follows: \\
\begin{definition}\label{def:stfm}
\textit{A Spatio-Temporal Foundation Model (STFM) is a large-scale neural network pre-trained on diverse spatio-temporal data sources, designed to generalize across multiple tasks by learning universal patterns of spatial and temporal dependencies.} \\
\end{definition} 
This description is intentionally broad to reflect the significant diversity within existing STFM research. We observe that the ``\textit{foundation-ness}" of foundation models is not a binary description, rather it exists on a spectrum with varying levels of demonstrated generalization capabilities. To begin to decipher this diversity, we pose the fundamental question: \textbf{\textit{what should a spatio-temporal foundation model be able to do?}} We address this question by identifying four main ways in which tasks can vary in the spatio-temporal context, leading to four forms of generalization by which to evaluate the capabilities of STFMs. In summary, these are as follows:

\begin{figure*}
    \centering
    \includegraphics[width=12cm,height=7cm]{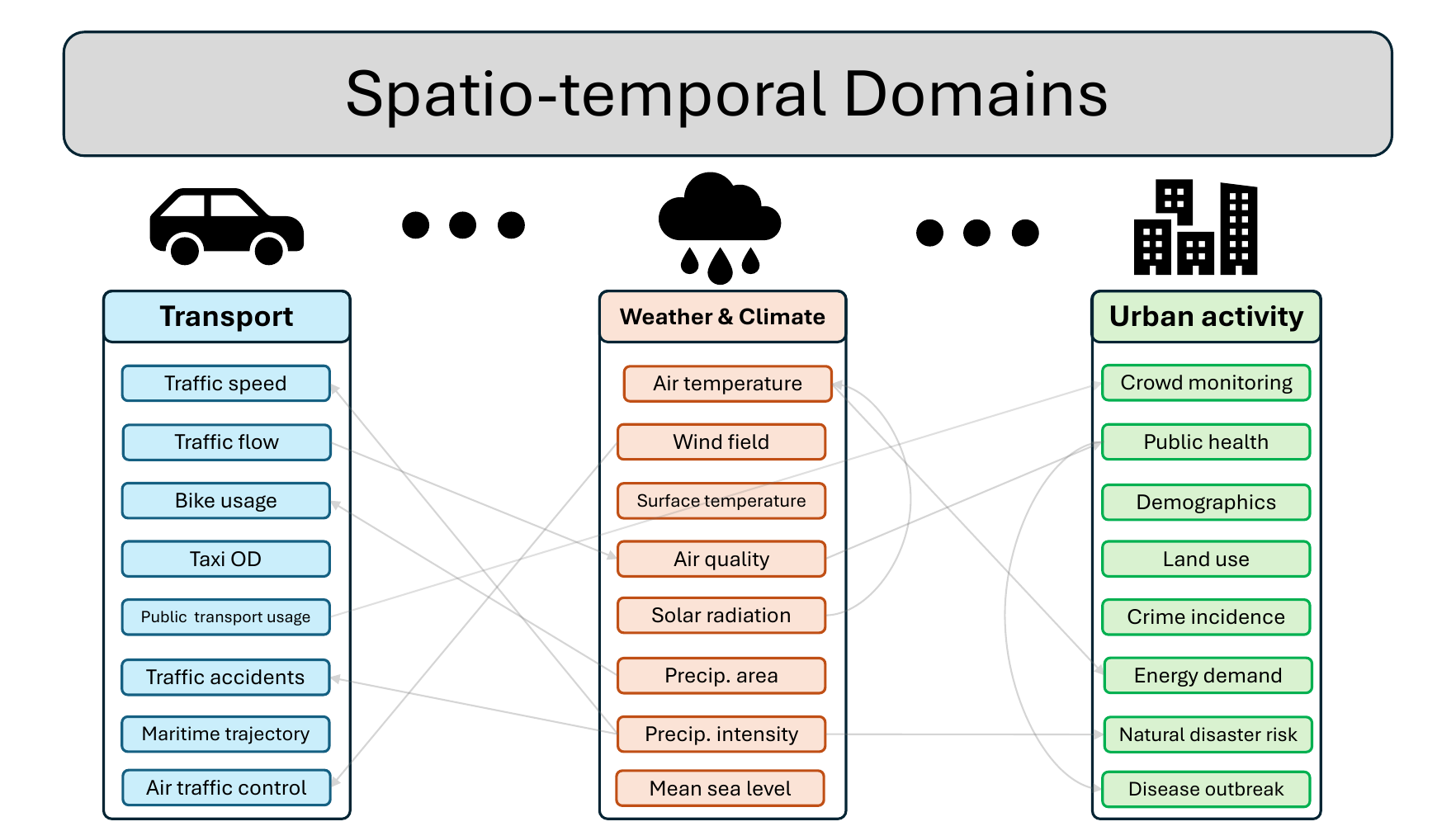}
    \caption{A selection of spatio-temporal domains that STFMs should be able to generalize across. Dashed lines indicate examples of potential for correlations or relationships in data patterns between different domains.}
    \label{fig:domains}
\end{figure*}

\begin{enumerate}
    \item \textbf{Domain} generalization: across different sources of data representing different physical systems and categories of applications.
    \item \textbf{Spatial} generalization: across different locations or regions in space.
    \item \textbf{Temporal} generalization: across different periods and intervals of time.
    \item \textbf{Scale} generalization: across different resolutions, frequencies or granularities of data.
\end{enumerate}
These generalization capabilities can be evaluated in two distinct ways:
\begin{itemize}
    \item \textbf{In-distribution (ID) generalization}: We can assess the model's ability to learn across multiple domains and distributions simultaneously during the pre-training stage by evaluating its performance on the variety of tasks seen during pre-training.

    \item \textbf{Out-of-distribution (OOD) generalization}: Alternatively, we can evaluate the model's ability to transfer to new tasks that were not observed during pre-training, even without prior exposure to their data distribution.
\end{itemize}

\subsection{Generalization Capabilities of STFMs}\label{sec:gc}

In the remainder of this section, we comprehensively detail the four types of generalization, and identify the key challenges in achieving them with current data and technological constraints.

\subsubsection{Domain Generalization}\label{sec:gc_domain}

As we have seen, ST data is prevalent in a highly diverse range of applications. Figure \ref{fig:domains} illustrates a small selection of these, organized by top-level categories or domains: transportation, weather \& climate and urban activity. Each of these categories encompasses a plethora of individual applications. For example, the transportation domain includes traffic flow measured by road network sensors, the time and location of traffic incidents, or the volume of passengers entering and exiting a public transport network at different stops. Within weather \& climate, there is a wide variety of different atmospheric variables, such as air temperature, precipitation, or the concentration of different pollutants. Given these diverse applications, the first type of generalization is across different domains of data.

\paragraph{Challenge}
Foundation models rely on the existence of shared patterns across different data sources or domains. For example, in language, the semantic meaning of words are generally consistent across various contexts, and sentences adhere to a common set of grammatical rules. In contrast, the rules that determine the distribution of ST data are highly application-dependent. For example, in a transportation network, applications like traffic flow and traffic incidents are likely to exhibit strong correlations because they both pertain to the same underlying physical system. Similarly, within weather \& climate studies, the concentration of pollutants in the air is often closely correlated with rainfall occurrence and intensity. In these cases, training a spatio-temporal foundation model (STFM) with data from both applications could offer mutual benefits by leveraging these shared patterns. However, the existence of shared patterns becomes much less clear between more disparate applications, such as traffic incidents and disease outbreaks. In such cases, it is uncertain whether training STFMs to simultaneously model both applications would enhance or degrade performance for either, a phenomenon known as negative transfer in the deep learning community. This highlights the challenges of developing STFMs that can generalize effectively across diverse applications. As discussed in Section \ref{sec:current_stfm}, current research typically focuses on more narrowly defined STFMs, often limited to a few applications within a single domain, rather than addressing cross-domain generalization.

\subsubsection{Spatial Generalization}\label{sec:gc_spatial}

The second type of generalization is across different locations in space. STFMs should not only be limited to applications from a limited selection of geographic spaces; it should learn from diverse environments and conditions simultaneously, and also be able to transfer to unseen places during inference.

\paragraph{Challenge} 

ST data can exhibit significant spatial heterogeneity. In other words, data patterns can vary significantly from one location to another, even within a single application. For example, in traffic flow, a model trained on traffic data from one city may struggle to generalize to another city with different road networks or traffic regulations. In pollutant concentration, a model trained on data from heavily urbanized areas may struggle to generalize to suburban or rural areas. This challenge is particularly acute in applications where existing dataset available for use in pre-training are heavily biased towards certain regions. For example, a disproportionate amount of the traffic datasets commonly used in existing studies are collected in a small number of major urban centers like Beijing, New York City, and London. This limitation increases the risk that STFMs are biased towards certain patterns that are most prominent in these and other similar cities, and failing to generalize to other regions with little or no representation in the training data.

\subsubsection{Temporal Generalization}\label{sec:gc_temporal}

STFMs should also be able to generalize across different periods of time. For example, it should perform well both during both day and night, on weekdays as well as weekends, and from one year to the next.

\paragraph{Challenge}

Spatio-temporal patterns are inherently dynamic, continuously evolving in complex ways. As a result, the patterns learned by models may lose relevance over time. These changes can occur gradually, as the characteristics of a given space evolve. For instance, general population growth in a city leads to a steady increase in traffic and public transport usage. Such gradual changes are typically easier to manage, as the slow pace of the data distribution shift provides an opportunity to adapt the model by re-training it with more recent data. In contrast, some changes are abrupt and drastic. These often result from interventions, such as the opening of a new attraction that drives a sudden surge in traffic, or from unforeseen events like natural disasters. Such shifts are far more challenging to accommodate, as their effects are complex and the distribution shift is akin to a step-change, creating a significant gap between the most recent historical data and the new reality. This limits the ability to retrain the model effectively in response to these sudden changes.

\subsubsection{Scale Generalization}

Spatio-temporal data spans a broad range of scales. Spatially, this can range from small-scale measurements, such as those at the meter level, to large-scale observations covering distances of hundreds of kilometers or more. Similarly, temporal scales vary from high-frequency observations, with timestamps in seconds or minutes, to low-frequency data collected over days, weeks, or months. STFMs must be capable of generalizing across these diverse spatial and temporal scales.

\paragraph{Challenge}
Spatio-temporal patterns can be highly \textit{scale-dependent}, i.e, patterns can differ significantly depending on the scale at which the data is observed.  This is particularly evident in weather applications, where a model trained on global weather patterns may not perform well on a fine-grained, region-level scale as individual regions have unique features and micro-climates which are less prominent in broader, global-level data. To overcome this, existing research in spatio-temporal data mining has explored hierarchical architectures, which are designed to capture patterns at multiple levels and across different scales.

\section{Current Research in STFMs}\label{sec:current_stfm}

In this section, we explore the current state of research on STFMs. As of this writing, we found no studies that comprehensively address the full range of spatial-temporal applications outlined earlier. Nevertheless, we identified six STFMs that align with the principles of foundation models as defined in Definition \ref{def:stfm}, three of which primarily focus on transportation-related applications and the other three focus on weather-related applications. We now provide a brief overview of these models.

\begin{table*}
    \centering
    \setlength{\tabcolsep}{6pt} 
    \renewcommand{\arraystretch}{1.3} 
    \caption{Generalization capabilities of existing STFMs.}
    \begin{tabular}{l|p{1.2cm}|p{5cm}c|p{3cm}c|cc|cc|c}
    \toprule
      \multicolumn{1}{c}{}  & \multicolumn{8}{c}{Generalization Capability} \\
     \multicolumn{1}{c}{STFM} & \multicolumn{1}{c}{Data Type} & \multicolumn{2}{c}{Domain} & \multicolumn{2}{c}{Spatial} & \multicolumn{2}{c}{Temporal} & \multicolumn{1}{c}{Scale} \\   
        \midrule
        UniST \cite{yuan2024unist} & Grid & Traffic speed, cellular usage, taxi demand, bike usage, crowd flow & \includegraphics[width=0.3cm]{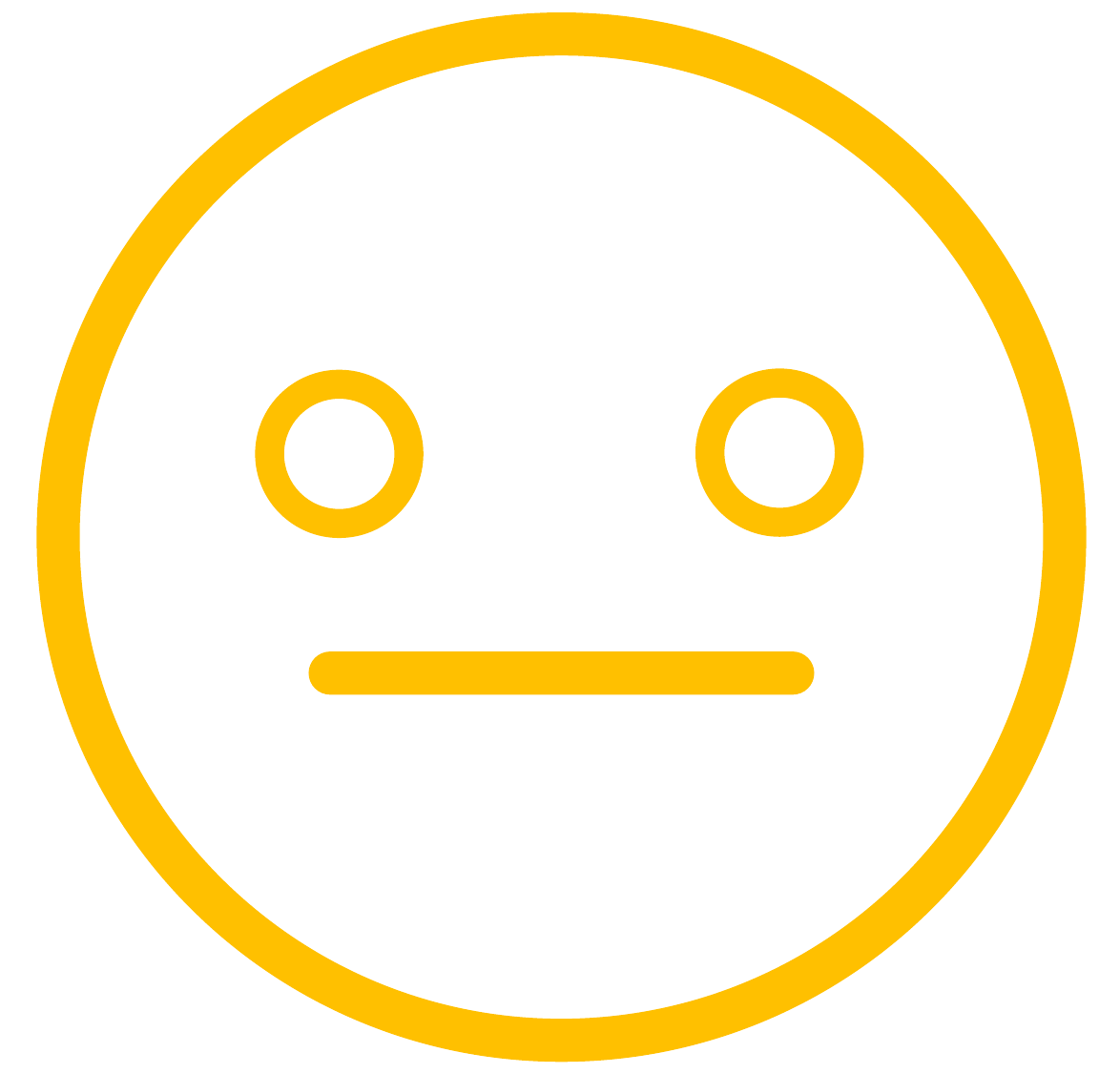} & Various US and Chinese cities & \includegraphics[width=0.3cm]{assets/mid.pdf} & 2013-2022 &  \includegraphics[width=0.3cm]{assets/mid.pdf} & \includegraphics[width=0.3cm]{assets/mid.pdf}  \\
        UrbanGPT \cite{li2024urbangpt} & Time-series &  Taxi usage, bike usage, crime & \includegraphics[width=0.3cm]{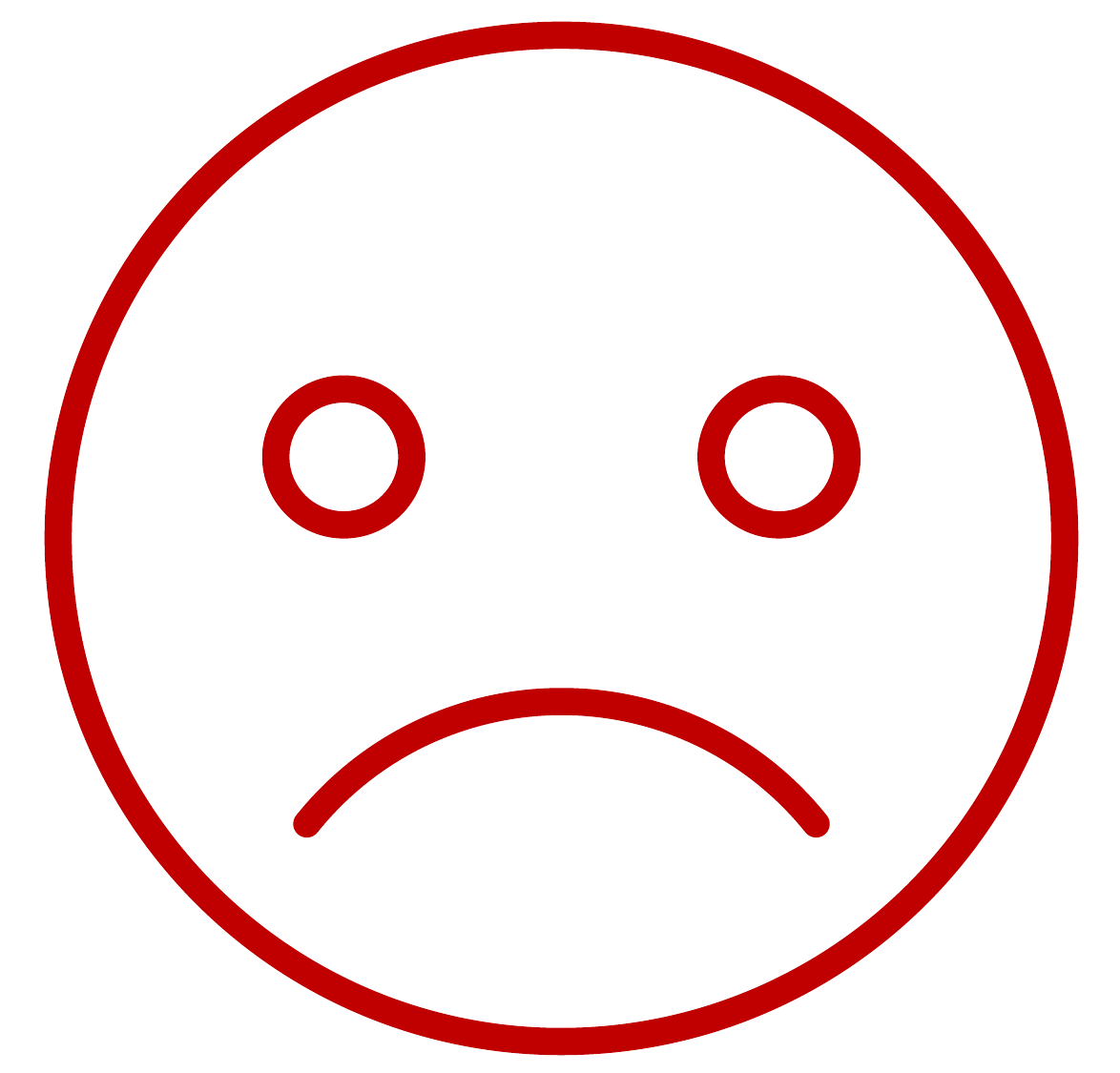} & New York City, Chicago & \includegraphics[width=0.3cm]{assets/bad.pdf} & 2017, 2021 & \includegraphics[width=0.3cm]{assets/mid.pdf} & \includegraphics[width=0.3cm]{assets/bad.pdf}  \\
        OpenCity \cite{li2024opencity} & Grid, sensor & Traffic flow, taxi demand, bike trajectory, traffic speed & \includegraphics[width=0.3cm]{assets/mid.pdf} & Various US and Chinese cities &  \includegraphics[width=0.3cm]{assets/mid.pdf} & 2013-2022 & \includegraphics[width=0.3cm]{assets/bad.pdf} & \includegraphics[width=0.3cm]{assets/mid.pdf}  \\
        W-MAE \cite{man2023w} & Grid & ERA5 reanalysis data &  \includegraphics[width=0.3cm]{assets/mid.pdf} & Global coverage & \includegraphics[width=0.3cm]{assets/mid.pdf} & 1979-2018 & \includegraphics[width=0.3cm]{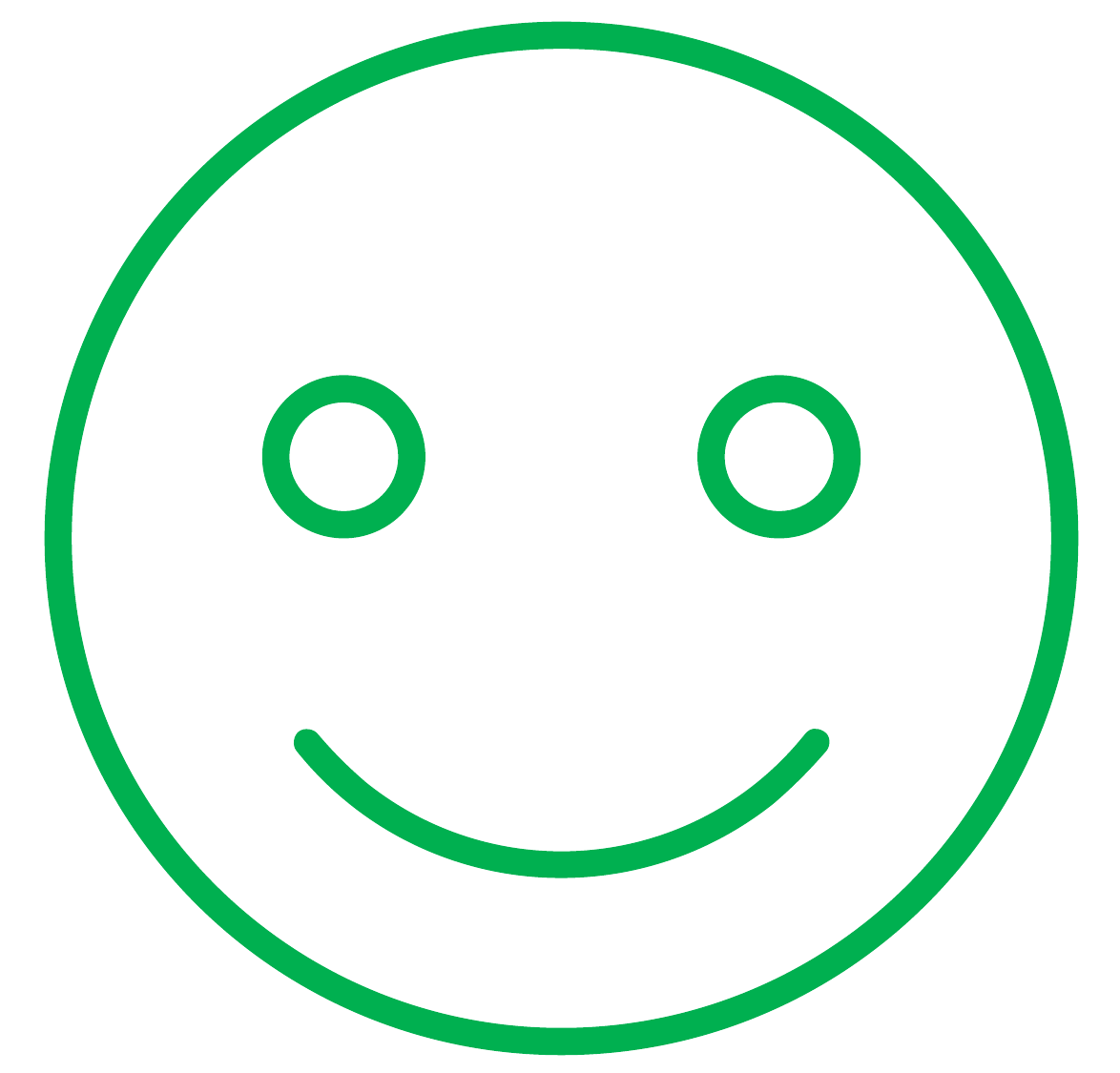} & \includegraphics[width=0.3cm]{assets/bad.pdf}\\
        Pangu-Weather \cite{bi2022pangu} & Grid & ERA5 reanalysis data & \includegraphics[width=0.3cm]{assets/mid.pdf} & Global coverage & \includegraphics[width=0.3cm]{assets/mid.pdf} & 1979-2021 & \includegraphics[width=0.3cm]{assets/good.pdf} & \includegraphics[width=0.3cm]{assets/bad.pdf} \\
        ClimaX \cite{nguyen2023climax} & Grid & CMIP6 simulations, ERA5 reanalysis data &  \includegraphics[width=0.3cm]{assets/good.pdf} & Global coverage & \includegraphics[width=0.3cm]{assets/mid.pdf} & 1850-2014 & \includegraphics[width=0.3cm]{assets/good.pdf} & \includegraphics[width=0.3cm]{assets/good.pdf}\\
    \bottomrule
    \end{tabular}
    \label{tab:stfm_generalization}
\end{table*}

\textbf{UniST} \cite{yuan2024unist}, \textbf{OpenCity} \cite{li2024opencity}, and \textbf{UrbanGPT} \cite{li2024urbangpt} are three STFMs designed for transportation applications. In particular, traffic prediction is the primary focus of both UniST and OpenCity, both of which adopt the transformer architecture to learn from large-scale, heterogeneous traffic datasets. UniST adopts a masked-patch modelling pre-training objective, where grid-based ST data is organised into ST patches, and the model is trained to reconstruct the data of masked patches given the unmasked patches. Various types of mask are used in pre-training to simulate different problem settings, and it also learns prompt memory pools to encode different forms of spatio-temporal knowledge for improved reconstruction. OpenCity instead combines a transformer with a graph neural network to learn distinct spatial and temporal patterns through separate attention layers and is trained with an absolute error loss function. On the other hand, UrbanGPT trains a spatio-temporal dependency encoder to a sequence of ST embedding tokens with the same dimensionality as the internal representation space of a large language model (LLM). The LLM is then used as the processing engine, given contextual information and a task description in the form of a natural language instruction, to predict the next tokens in the sequence, which are converted back into ST data through a regression layer.

\textbf{ClimaX} \cite{nguyen2023climax}, \textbf{Pangu-Weather} \cite{bi2022pangu}, and \textbf{W-MAE} \cite{man2023w} are three weather-focused STFMs. In this context, there are two massive-scale broad datasets for weather science: CMIP6 climate simulations and \cite{meehl2000coupled}, and ERA5 reanalysis data \cite{store2021era5, hersbach2020era5}. Both of these datasets offer global coverage of a wide variety of atmospheric variables over extremely long temporal horizons. The primary focus of these STFMs is the ability to forecast the future state of several of these variables simultaneously. ClimaX is based on a vision transformer, and it tokenizes  each variable independently in order to flexibly handle different sets of input variables. To reduce memory complexity, it aggregates these embeddings to a single vector representation for each spatial unit. The model is trained to minimize the latitude-weighted prediction error. Pangu-Weather designs a 3D Earth Specific Transformer (3DEST) that merges surface-level and upper-air variables into cubic data patches, and employs a encoder-decoder architecture to learn their patterns, also trained with an error-based loss function. On the other hand, W-MAE employs a masked auto-encoder pre-training objective with a ViT backbone architecture for multi-variable weather forecasting. It is important to note that all three weather foundation models use a single time-stamp of historical observations as input to predict the future state of the atmosphere at a variety of lead times, which means they do not explicitly model temporal dependencies in order to make predictions.

Table \ref{tab:stfm_generalization} shows our qualitative assessment of the generalization capabilities of these STFMs, based on their demonstrated performance in their original work. We considered two main factors when determining our assessment: (1) \textit{does the choice of data used in pre-training contribute towards this form of generalization?} and (2) \textit{is there an effort to experimentally evaluate or demonstrate this form of generalization capability?} We denote a positive assessment (i.e. significant evidence) as \includegraphics[width=0.3cm]{assets/good.pdf}, a neutral assessment (i.e. some but insufficient evidence) as \includegraphics[width=0.3cm]{assets/mid.pdf}, and a negative assessment (i.e. little to no evidence) as \includegraphics[width=0.3cm]{assets/bad.pdf}. We emphasize that our assessment \textit{should not be interpreted as a reflection or comparison of their performance}, rather a study of the extent to which current research aligns with the principles of foundation models.

\subsubsection{Domain Generalization}

As mentioned earlier, our immediate observation is the significant fragmentation between high-level applications. UniST, UrbanGPT and OpenCity are exclusively focused on transportation applications. Both UniST and OpenCity are both trained or evaluated on 21 datasets in total, but the majority of these relate to traffic speed and traffic flow, with only a few other datasets for other applications like bike usage, taxi demand and cellular usage. Both models are evaluated in both ID generalization by training on multiple applications simultaneously, as well as OOD generalization by leaving certain datasets out of the training distribution. On the other hand, UrbanGPT uses only four datasets: taxi, bike, and crime datasets from New York City and a taxi dataset from Chicago. The model is tested in ID generalization across the three training datasets, but not OOD generalization. 
 
As for weather-focused STFMS, there are a greater range of features in the form of different atmospheric variables, however all three models evaluate perdiction performance for only 4 of them. Pangu-Weather is also evaluated with a tropical cyclone tracking task, which is derived from the local minmum of mean sea level pressure.

\subsubsection{Spatial Generalization}

The spatial coverage of publicly available traffic datasets are limited to only a few major urban centers. The datasets used in both pre-training and evaluation of UniST and OpenCity come from various cities in the United States and China, and they are evaluated in their generalization to unseen regions of these cities. UrbanGPT is exclusively trained on data from New York City, and evaluated in its generalization to unseen regions with the city, as well as to a new city (Chicago). As these geographic locations are very limited, it is difficult to assess their generalization capabilities to locations which are highly dissimilar to the training set. 

As mentioned, the two weather datasets which are used in both pre-training and evaluating the weather foundation models are global in their spatial coverage. ClimaX is trained on CMIP6 \cite{meehl2000coupled} and fine-tuned and tested on ERA5 \cite{store2021era5, hersbach2020era5}, while Pangu-Weather and W-MAE are trained and evaluated on different years of data from ERA5. Evaluation of all models is performed across the entire globe in aggregate, though ClimaX also performs regional forecasting for the North America region specifically. As both training and inference is performed over the entire globe, this does not demonstrate generalization from one seen regions to unseen regions, and also lacks fine-grained analysis of how errors are spatially distributed across different regions.

\subsubsection{Temporal Generalization}

Most of the traffic datasets from China are recorded within the same one month period in March-April 2022, at intervals at 5 minutes. The other datasets are slightly more varied in time period and total length over the past decade. UrbanGPT is evaluated in its long-term forecasting capabilities by training the model on 2017 data and testing it on 2021 data.

The weather datasets generally contain much greater timespans, enabling training and evaluation over greater temporal ranges. ClimaX is trained on data from 1850 to 2014 from CMIP6, and tested on ERA5 from 1979 to 2018. Pangu-Weather is trained on 38 years of data from ERA5 (1979-2017), validated in 2019 and tested in 2018 and 2020-2021 data, while W-MAE trains two separate models: one with 2 years and one with 37 years of data.

\subsubsection{Scale Generalization}

Most of the traffic datasets are recorded over 5 to 30 minute intervals. UniST performs short-scale experiments with 6 time steps and long-scale experiments with 64 time steps in both input and target output size. UrbanGPT only considers one temporal scale of 12 time-steps. 

Generalization across spatial scales is most relevant to the weather-related applications. Spatially, Pangu-Weather considers only one spatial resolution; $0.25\degree \times 0.25\degree$, which corresponds to approximately 28km $\times$ 28km cells. Comparatively, ClimaX evaluates performance with $5.625\degree \times 5.625\degree$ cells and $1.40625\degree \times 1.40625\degree$ cells. It also evaluates the model's ability to perform climate down-scaling from 5.625$\degree$ to 1.40625$\degree$, as well as regional forecasting over North America only. However, these are all relatively large-scale and coarse-grained resolutions, which fails to capture the local patterns which are most useful for local-scale prediction. Temporally, Pangu-Weather trains four separate models with different lead times (1, 3, 6 and 24 hours), and aggregates these models for any-time forecasting. This is in order to reduce the impact of error propagation in forecasting, especially at longer lead times, however it is also fundamentally at odds with the intended purpose of a foundation model. On the other hand, ClimaX randomizes lead times between 6 hours and 168 hours (1 week) in pre-training, and various other lead times for evaluation. Furthermore, they consider various forecasting tasks, including sub-seasonal prediction and climate projection with much longer temporal horizons. \\
\newline 
\noindent \textbf{Overview}: We make the following general observations from our assessment of existing STFM research. 

\begin{itemize}
    \item Existing STFMs are fragmented into transportation foundation models and weather foundation models with no overlap in applications or evaluations. Within each of these categories, the transportation foundation models are primarily focused on the task of traffic prediction, with a few datasets from applications like taxi demand, bike usage and crowd levels either included in the training sets or used to measure OOD generalization in the test set. On the other hand, weather foundation models often incorporiate a wide variety of atmospheric variables in their input variables, but are mostly evaluated in their prediction of only a small subset of these variables during inference.
    \item There are numerous datasets used to train and evaluate transportation foundation models, however they are critically limited in spatial coverage to just a small selection of major urban centers, almost entirely limited to cities located in the USA and China.
    \item Weather foundation models are exclusively trained on one or both of two massive-scale weather datasets, CMIP6 and ERA5, which offer global spatial coverage over extremely long temporal horizons. However, they are both limited to coarse granularities which limits their ability to generalize to applications which require more fine-grained analysis over smaller spatial regions.
\end{itemize}

\section{Opportunities}

In this section, we outline a range of potential avenues and directions for future research, both to address the limitations identified in the previous section, and also to unlock new performance enhancements and capabilities.

\subsection{Unified Architectures}

As discussed in Section \ref{sec:data}, ST data exists in a variety of types and formats. However, existing STFMs predominantly focus on rasterized grid-based data, often requiring data in other formats to be transformed into a structured grid, which can introduce significant information loss and obscure fine-grained spatiotemporal relationships. Ideally, STFMs should be designed to handle heterogeneous ST data natively, rather than relying on lossy pre-processing techniques. A number of recent works have started to explore the integration of more flexible ST graph-based data representations \cite{yuan2024foundation, yuan2024urbandit}. However, these approaches are still in their early stages and often treat graph-based structures as a secondary input, rather than a primary modeling component. This can lead to suboptimal representations that fail to fully capture the complex spatial dependencies and topological relationships inherent in many ST datasets. A more principled approach is needed to develop STFM architectures that seamlessly integrate structured and unstructured data.

Another limitation is the reliance on transformer-based architectures. While transformers have demonstrated strong performance in many ST applications, their quadratic complexity due to the self-attention mechanism presents a significant scalability challenge, particularly as ST datasets grow in size and complexity. This issue is exacerbated when dealing with high-resolution spatial data, long temporal sequences, or multi-modal inputs that require modeling diverse interactions across time and space. Addressing these challenges necessitates novel architectural innovations, such as sparse attention mechanisms, hierarchical modeling techniques, and hybrid approaches that combine the strengths of transformers with more computationally efficient architectures like convolutional neural networks (CNNs), recurrent neural networks (RNNs), and graph neural networks (GNNs).

Furthermore, there is a pressing need for more standardized benchmarks and evaluation protocols for STFMs. Existing studies use disparate datasets and evaluation settings. For example, UniST evaluates short-term and long-term prediction performance using a temporal horizon of 6 time-steps and 64 time-steps respectively, whereas OpenCity uses 1 day of observations, which means the number of time-steps varies between datasets depending on their frequency. Similarly, inconsistencies in spatial resolutions, training and test applications, evaluation metrics, and ID vs OOD generalization settings make it challenging to compare performance between models. Establishing comprehensive ST benchmarks that encompass diverse data types and downstream tasks would provide a clearer picture of model capabilities and drive further advancements.

\subsection{Cross-domain Synergies}

Section \ref{sec:gc_domain} explored the complex, under-utilized relationships between spatio-temporal (ST) sequences from different applications or sources.  Current ST models often treat each application in isolation, neglecting the rich information shared across domains. For example, based on our understanding of how infectious diseases spread through close contact, we can infer that patterns of human mobility are intricately linked to the spread of transmissible diseases within a population. For this reason, a model that accurately captures traffic flow and population movement can offer valuable insights for predicting disease outbreaks, informing public health interventions, and ultimately mitigating their impact. By training STFMs to recognize and leverage these cross-domain correlations, we can unlock significant advancements in individual applications. A model trained on diverse datasets, encompassing weather patterns, traffic data, social media activity, and epidemiological information, could learn more robust and generalizable representations of spatio-temporal dynamics, leading to improved performance in each individual domain.

Crucially, the relationships between different domains are often directional. Weather patterns influence travel behavior, making accurate weather forecasting a valuable input for predicting traffic flow. However, the reverse is not true; traffic congestion does not significantly impact the weather.  Therefore, incorporating prior knowledge about these directional relationships, such as physical laws and constraints, is essential for building effective STFMs.  For instance, incorporating known relationships between temperature, humidity, and evapotranspiration could enhance agricultural yield prediction models.

When the nature of the relationships between features is unknown, recent advances in causal learning offer powerful tools for discovery. Causal inference techniques can help disentangle correlation from causation, revealing the true drivers of spatio-temporal phenomena. This is particularly important for understanding complex systems where multiple factors interact in intricate ways.  While established research exists on discovering and inferring causal relationships in spatio-temporal data \cite{liu2011discovering, chu2014causal, zhu2017pg}, the rise of spatio-temporal neural networks has opened new avenues for exploration. Recent work has begun to integrate causal reasoning into these models, including generative approaches \cite{zhao2023generative}, graph neural networks for causal discovery \cite{wang2022causalgnn}, methods for deciphering causal relationships from observational data \cite{xia2024deciphering}, and frameworks for building causally aware ST models \cite{christiansen2022toward}. Future research should focus on developing novel methods for incorporating causal knowledge directly into STFMs, enabling them to learn more interpretable, robust, and generalizable representations of the world.  This includes exploring how to represent causal relationships within the model architecture, how to train models to respect causal constraints, and how to leverage causal information for improved prediction and forecasting.  Furthermore, investigating how to handle noisy and incomplete data in causal discovery within a spatio-temporal context is crucial for real-world applications.

\subsection{Multi-modal Training}

Information in the physical world is inherently multi-modal, encompassing diverse data sources such as numerical time-series, remote sensing imagery, textual descriptions, audio signals, and more. Integrating multi-modal learning into spatio-temporal foundation models (STFMs) has the potential to significantly enhance their ability to capture complex dependencies and improve their robustness across real-world challenges. By leveraging complementary information from different modalities, STFMs can achieve a more holistic understanding of spatio-temporal phenomena, reducing uncertainty and improving predictive performance.

For example, predicting crop yields could benefit from fusing satellite imagery (providing information on vegetation health and land cover), weather time-series data (temperature, rainfall), and soil sensor readings (moisture, nutrient levels).  Similarly, for urban traffic forecasting, combining traffic camera footage (visual flow), GPS data from vehicles (speed and location), and social media posts (reports of accidents or road closures) could lead to more accurate and robust predictions.  In disaster response, multi-modal data like satellite imagery (damage assessment), weather radar (storm tracking), and social media reports (citizen accounts of the situation) can provide a more comprehensive understanding of the event and improve the efficiency of relief efforts. Maps serve as a particularly crucial source of structured knowledge and constraints for learning from spatial-temporal data. As a structured representation of the contextual characteristics of the environment, maps can help to provide hard constraints, such as physical barriers that restrict movement (e.g., rivers, buildings, or restricted zones), and soft constraints, such as land-use characteristics that influence event likelihoods (e.g., residential versus commercial areas). They also serve as a regularization mechanism, guiding model learning through spatial smoothing techniques and topology-aware constraints, such as enforcing connectivity in road network-based predictions using graph neural networks.

There have recently been some initial strides towards curating large-scale, multi-modal datasets for spatio-temporal applications. For example, the Terra dataset \cite{chenterra} encompasses 45 years’ worth of hourly time-series data from over 6 million grid areas worldwide, covering various meteorological variables while incorporating multi-modal information such as explanatory text, geo-imaging, and satellite imagery. However, these efforts remain scarce and largely limited to a few applications and modalities.

One major challenge in multi-modal ST learning is the alignment and fusion of heterogeneous data sources, which often have vastly different temporal resolutions, spatial granularities, and noise characteristics. For instance, while satellite imagery may be available at a high spatial resolution but captured at irregular intervals, time-series sensor data may provide continuous readings but at a coarser spatial resolution. Effective STFMs must develop novel alignment techniques to efficiently integrate such disparate data sources without introducing artifacts or losing critical information.

\subsection{Adaptation to Distribution Shift}

Foundation models are trained with large-scale data, across diverse sources and domains, which exposes them to a wide range of data distributions. Nevertheless, during inference, they remain heavily reliant on the statistical properties of their training data. If the test data significantly deviates from the training distribution, the model's performance can drop sharply. In practice, the physical environment is constantly evolving which guarantees some form of distribution shift in data over time, e.g. temporal changes like urban development or changes in regional policies. Moreover, the particularly high complexity and dimensionality of spatio-temporal data increases the likelihood of encountering unseen patterns or out-of-distribution (OOD) scenarios in practice. 

Foundation model adaptation \cite{liu2024few,chambon2022adapting,khattak2023self} offers a promising approach to address this challenge and has received significant research attention in order to boost their performance in individual vision and language tasks, particularly involving data outside the original training distribution. Several adaptation strategies have shown promise for addressing OOD challenges. For example, domain-adversarial training encourages the model to learn representations invariant to the specific domain of the training data. By minimizing the discrepancy between feature distributions of different domains, the model becomes more robust to domain shifts and generalizes better to unseen data. In the ST context, this could involve training the model to be invariant to specific geographic regions or time periods, enhancing adaptability to new locations or future conditions. Meta-learning aims to train models that can rapidly adapt to new tasks or distributions with only a few examples. The model learns a set of parameters well-suited for fast adaptation, enabling quick adjustments to new ST patterns with limited data. This is particularly relevant for ST applications where labeled data in new domains might be scarce. Test-time adaptation adjusts the model's parameters during inference using the unlabeled test data itself, proving effective for handling gradual distribution shifts over time. Techniques like self-training and entropy minimization can be used to adapt the model to the characteristics of the current test batch, improving performance on unseen data.

\section{Conclusion}

Spatio-temporal foundation models offer an exciting development with great potential to improve performance in existing tasks, as well as unlock capabilities in new tasks related to spatio-temporal data. While STFMs have shown significant promise in capturing the intricate relationships between space and time, their ability to generalize to unseen spatial regions, time periods, and novel downstream tasks remains a critical challenge. We highlighted key issues such as spatial variability, temporal dynamics, data distribution shifts, and scale-dependent patterns that hinder effective generalization. As spatio-temporal models continue to evolve, future research should focus on developing methods to enhance their adaptability to unseen data distributions and dynamic environments. By integrating more robust mechanisms for handling cross-domain correlations and incorporating fine-grained adaptation techniques, we can unlock the full potential of STFMs for a wide range of real-world applications. Through continued innovation and refinement, spatio-temporal models have the potential to advance our understanding of complex systems and enable more accurate predictions, better decision-making, and improved outcomes across domains with important real-world implications.

\bibliographystyle{plain}
\bibliography{ref}
\end{document}